\documentclass{article}

    \PassOptionsToPackage{numbers}{natbib}

    \usepackage[preprint]{neurips_2025}

\usepackage[utf8]{inputenc} 
\usepackage[T1]{fontenc}    
\usepackage[pagebackref, breaklinks=true, colorlinks,citecolor=green, linkcolor=red, bookmarks=false]{hyperref}
\usepackage{url}            
\usepackage{booktabs}       
\usepackage{amsfonts}       
\usepackage{nicefrac}       
\usepackage{microtype}      
\usepackage{amssymb}
\usepackage{graphicx}
\usepackage{adjustbox}
\usepackage{booktabs} 
\usepackage{color}
\usepackage{amsmath}
\usepackage{caption}
\usepackage{multirow}
\usepackage{wrapfig}
\usepackage{graphicx}
\usepackage{subcaption}
\usepackage[dvipsnames, table]{xcolor}

\definecolor{ceiling}{RGB}{214, 8, 40}   %
\definecolor{floor}{RGB}{43, 160, 4}     %
\definecolor{wall}{RGB}{158, 216, 229}  %
\definecolor{window}{RGB}{114, 158, 206}  %
\definecolor{chair}{RGB}{204, 204, 91}   %
\definecolor{bed}{RGB}{255, 186, 119}  %
\definecolor{sofa}{RGB}{147, 102, 188}  %
\definecolor{table}{RGB}{30, 119, 181}   %
\definecolor{tvs}{RGB}{160, 188, 33}   %
\definecolor{furniture}{RGB}{255, 127, 12}  %
\definecolor{objects}{RGB}{196, 175, 214} %
\definecolor{redarrows}{RGB}{192, 0, 0} %

\title{Occupancy World Model for Robots}

\author{
    \vspace{1mm}
    Zhang Zhang$^{1,2,*}$,
    Qiang Zhang$^{1,3,*}$,
    Wei Cui$^{1,*}$,
    \\
    \vspace{1mm}
    Shuai Shi$^{1}$,
    Yijie Guo$^{1}$,
    Gang Han$^{1}$,
    Wen Zhao$^{1}$,
    \\
    \vspace{1mm}
    Jingkai Sun$^{1,3}$,
    Jiahang Cao$^{1,3}$,
    Jiaxu Wang$^{1,3}$,
    Hao Cheng$^{1,3}$,
    \\
    \vspace{1mm}
    Xiaozhu Ju$^{1}$,
    Zhengping Che$^{1}$,
    Renjing Xu$^{3}$,
    Jian Tang$^{1,\dagger}$\\ \vspace{1mm}
    $^{1}$ Beijing Innovation Center of Humanoid Robotics \\ \vspace{1mm}
    $^{2}$ Beijing Institute of Technology \\ \vspace{1mm}
    $^{3}$ Hong Kong University of Science and Technology (Guangzhou) \\ \vspace{1mm}
    \footnotesize{$^*$ Contributed equally.}
    \footnotesize{$^\dagger$ Corresponding author.}
}

\begin{document}
    \renewcommand\twocolumn[1][]{#1}%
    \maketitle
    \vspace{-10mm}
    \begin{center}
        \centering
        \includegraphics[width=\linewidth]{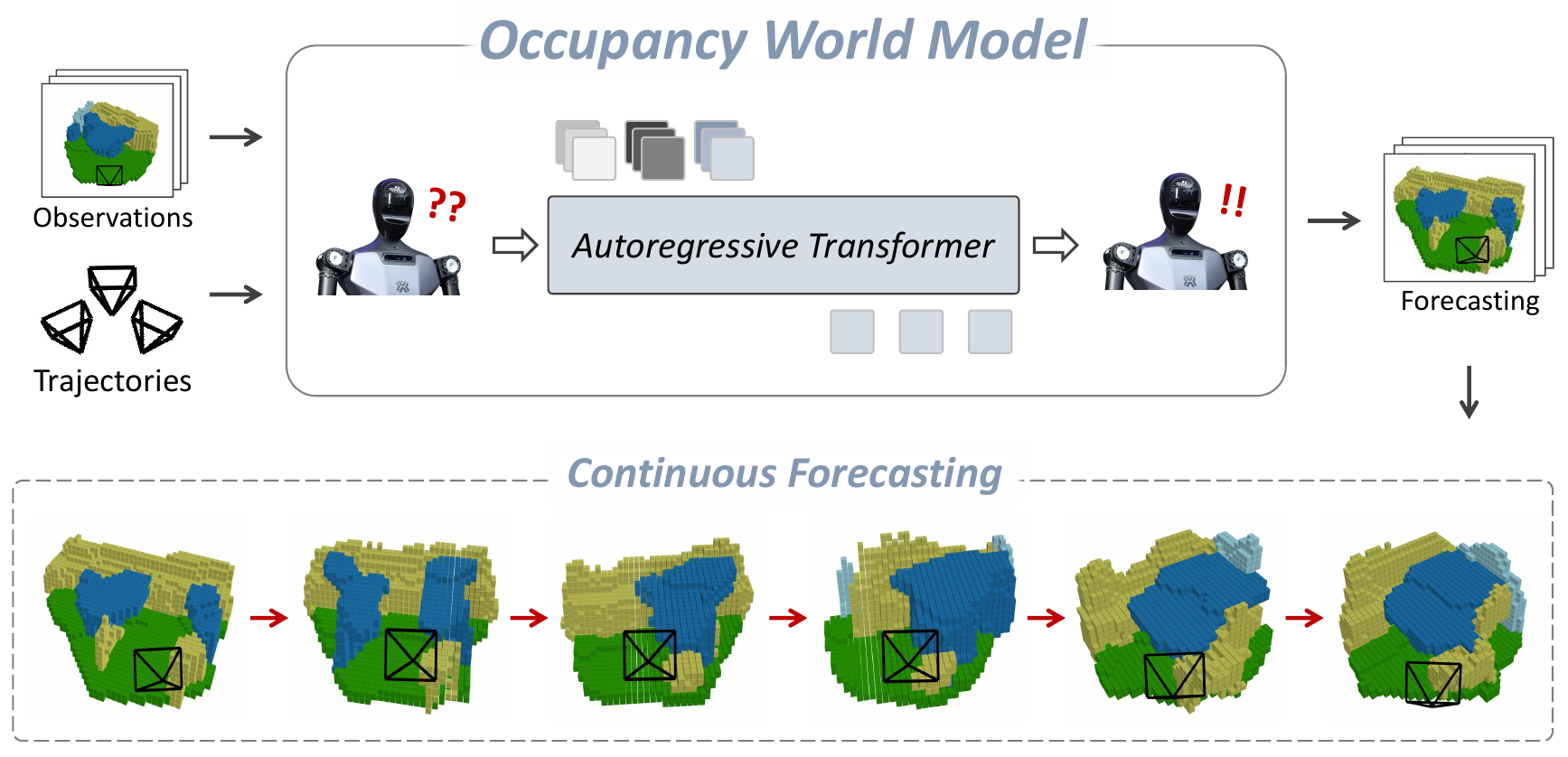}
        \vspace{-2mm}
        \captionof{figure}{Given previous 3D occupancy observations and next-step trajectories, our occupancy World model based on the combined spatio-temporal receptive field and guided autoregressive transformer can forecast scene evolutions for robots' decision and exploration.}
    \label{fig_1}
    \end{center}

\begin{abstract}
    Understanding and forecasting the scene evolutions deeply affect the exploration and decision of embodied agents. While traditional methods simulate scene evolutions through trajectory prediction of potential instances, current works use the occupancy world model as a generative framework for describing fine-grained overall scene dynamics. However, existing methods cluster on the outdoor structured road scenes, while ignoring the exploration of forecasting 3D occupancy scene evolutions for robots in indoor scenes. In this work, we explore a new framework for learning the scene evolutions of observed fine-grained occupancy and propose an occupancy world model based on the combined spatio-temporal receptive field and guided autoregressive transformer to forecast the scene evolutions, called RoboOccWorld. We propose the Conditional Causal State Attention (CCSA), which utilizes camera poses of next state as conditions to guide the autoregressive transformer to adapt and understand the indoor robotics scenarios. In order to effectively exploit the spatio-temporal cues from historical observations, Hybrid Spatio-Temporal Aggregation (HSTA) is proposed to obtain the combined spatio-temporal receptive field based on multi-scale spatio-temporal windows. In addition, we restructure the OccWorld-ScanNet benchmark based on local annotations to facilitate the evaluation of the indoor 3D occupancy scene evolution prediction task. Experimental results demonstrate that our RoboOccWorld outperforms state-of-the-art methods in indoor 3D occupancy scene evolution prediction task. The code will be released soon.
\end{abstract}

\section{Introduction}

    With the remarkable development of embodied intelligence \cite{qi2019deep, liu2024aligning, qi2020imvotenet, couprie2013indoor, zhang2025humanoidpano, liu2024volumetric, wang2024embodiedscan, xu2024embodiedsam}, enabling robots to explore in unknown environments has received widespread attention in recent years. Robots based on the perception and prediction could make decisions ahead of time and act better when exploring indoor scenes and executing downstream tasks. While most conventional methods \cite{salzmann2020trajectron++, giuliari2021transformer, ngiam2021scene, zhou2022hivt, tang2024hpnet} follow the pipeline which predicts the trajectory of potential instances to plan ahead and make better judgments about what comes next, existing works \cite{zheng2024occworld, wang2024occsora} use the occupancy world model as a generative framework for describing fine-grained overall scene dynamics and make remarkable results. However, current methods \cite{zheng2024occworld, zhang2024efficient, wei2024occllama, yang2025driving, yan2024renderworld, xu2025occ, gu2024dome} focus on outdoor scenes, while ignoring the exploration of 3D occupancy scene evolutions for robots in indoor scenes. To bridge the gap between existing research and practical scenarios, we restructure the OccWorld-ScanNet benchmark based on local annotations and formulate an indoor 3D occupancy scene evolution prediction task to evaluate the next state occupancy prediction and autoregressive occupancy prediction. 
    
    However, when we test the method \cite{zheng2024occworld} which uses the GPT-like autoregressive transformer architecture to predict the occupancy of the next state only based on historical frame in indoor scenes, the predictions are difficult to be optimal. The sensors of autonomous vehicles generally remain fixed and their movements are restricted on structured roads, which makes the frame-to-frame variations of the history information small and further preserves the spatio-temporal regularity of the history frames, but that won't work in the indoor scene for robots. Due to the flexible perspective changes and unfettered movement of robots, there is no tight correlation between the next frame and the current one that observed. As shown in Figure \ref{fig_2}, we show three different camera pose sequences and their corresponding observation sequences in same scene. What can be seen is that the robot does not follow the single direction in its exploration, and its next camera pose is not tightly linked to the current state. Compared to outdoor data, the diversity of camera positions and angles is greatly increased in indoor scenes, which not only makes the historical information no longer spatio-temporal regular among themselves, but also makes autoregressive prediction difficult. To address this, we propose the Conditional Causal State Attention (CCSA), which treats the next camera pose as the condition to guide the autoregressive transformer. The proposed module utilizes visual observations and potential camera poses to predict future fine-grained occupancy evolutions, allowing the network to understand the spatio-temporal connections.

    \begin{wrapfigure}{l}{0.5\textwidth}
    \centering
    \includegraphics[width=\linewidth]{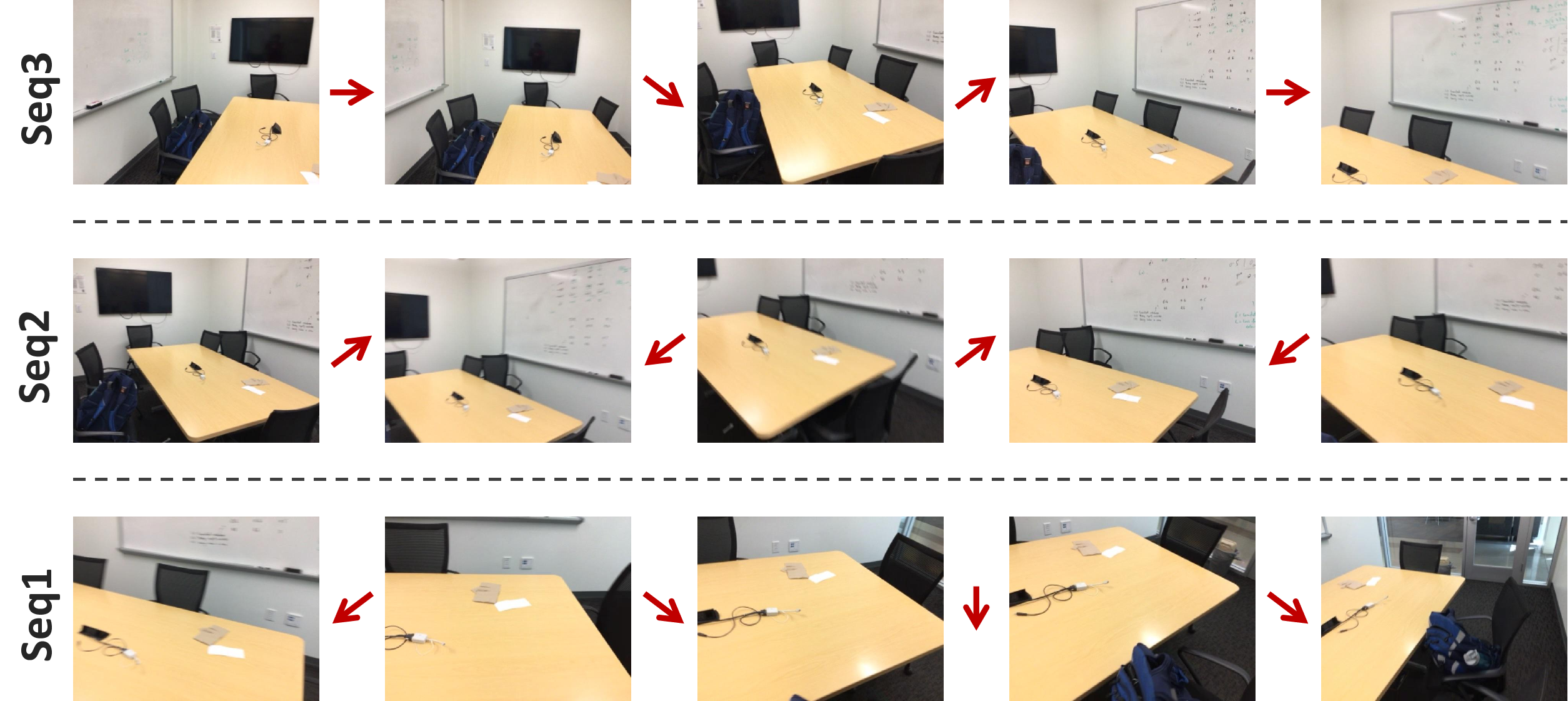}
    \caption{\textbf{The indoor monocular images of different sequences.} The {\textcolor{redarrows}{red arrows}} corresponding to the images represent the direction of camera movement.}
    \label{fig_2}
    \end{wrapfigure}
    
    Further, existing method \cite{zheng2024occworld} utilizes spatial aggregation and temporal causal attention to obtain the spatio-temporal interaction. As shown in Figure \ref{fig_3} (a), spatial aggregation first acquires spatial information within frames, and subsequently temporal causal attention performs grid-to-grid interactions between frames. However, the spatial and temporal interactions are performed separately, resulting in insufficient spatio-temporal receptive fields to effectively exploit the spatio-temporal cues of historical frames. Inspired by this, we propose the Hybrid Spatio-Temporal Aggregation (HSTA), which obtains the combined spatio-temporal receptive field by introducing the multi-scale spatio-temporal windows. Specifically, we capture the short-term  fine-grained scene evolutions and the long-term spatio-temporal coherence based on the short-term window and the long-term window, respectively. As shown in Figure \ref{fig_3} (b), the different colored blocks represent different windows. The spatio-temporal clues of the historical scenes are retrieved through the effective feature aggregation from spatial and temporal windows of different sizes.

    \begin{figure*}[h]
    \centering
        \includegraphics[width=1\textwidth]{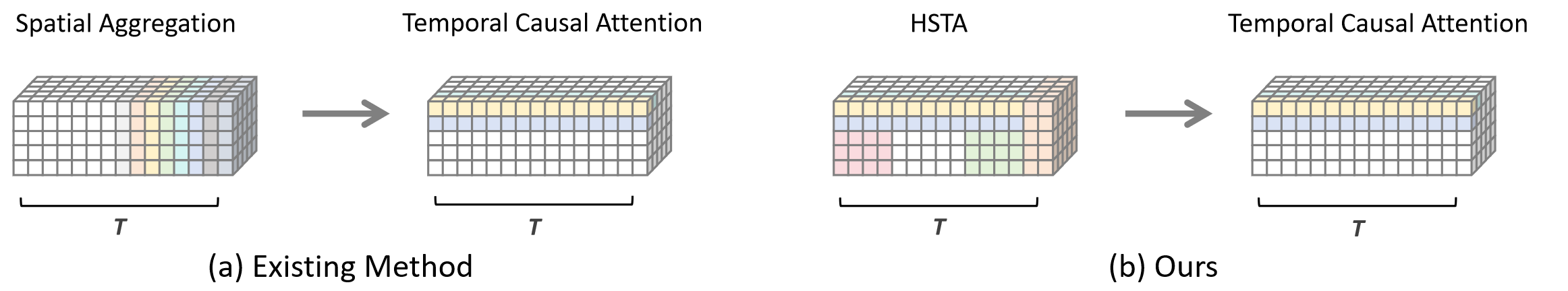}
        \caption{\textbf{The comparison between existing methods and our method on spatio-temporal interaction.} The \textcolor{orange}{di}\textcolor{RawSienna}{ff}\textcolor{pink}{er}\textcolor{ForestGreen}{ent} \textcolor{SkyBlue}{co}\textcolor{Orchid}{lo}\textcolor{RedOrange}{red} \textcolor{RoyalBlue}{blo}\textcolor{Gray}{cks} represent different windows and $T$ is the time dimension.}
        \vspace{-2mm}
        \label{fig_3}
    \end{figure*}

    Our experimental results show that RoboOccWorld outperforms state-of-the-art methods in indoor 3D occupancy scene evolution prediction task by a substantial margin in both next state occupancy prediction and autoregressive occupancy prediction tasks. Our main contributions are summarized as follows:
    \begin{itemize}
    
    \item Conditional Causal State Attention (CCSA) is proposed, which utilizes visual observations and potential camera poses to guide the generative occupancy world model to make reasonable occupancy forecasts in indoor scenes for robots.
    
    \item  We utilize the Hybrid Spatio-Temporal Aggregation (HSTA) to obtain the combined spatio-temporal receptive field based on multi-scale spatio-temporal windows and the spatio-temporal cues from historical observations.

    \item We restructure the OccWorld-ScanNet benchmark to evaluate the next state occupancy prediction and autoregressive occupancy prediction. Experimental results show that RoboOccWorld outperforms state-of-the-art methods in indoor scene evolution prediction with occupancy representation.
    
    \end{itemize}
   
\section{Related Work}

\subsection{3D Occupancy Prediction}
    3D occupancy \cite{cao2022monoscene, huang2023tri, wei2023surroundocc} is now widely used as a generalized 3D representation for tasks such as end-to-end autonomous driving \cite{hu2023planning, tian2023occ3d, wang2023openoccupancy} and outdoor world model simulators \cite{zheng2024occworld, wang2024occsora}. It aims to predict whether the voxels in 3D space are occupied or not, and the semantics of the occupied voxels. Due to its comprehensive and flexible description of the 3D space, it has received increasing attention and has been expanded into various scenarios (e.g. indoor scenarios) \cite{silberman2012indoor, dai2017scannet, dai2018scancomplete, yu2024monocular}. Recent methods \cite{wu2024embodiedocc, zhang2025roboocc} have explored 3D occupancy-based scene understanding and continuous perception in indoor scenes for robots with remarkable results. However, existing methods focus only on acquiring indoor 3D semantic occupancy and ignore its temporal evolutions, which is crucial for exploration and decision of robots. In this paper, we explore a new framework for learning the scene evolutions of observed fine-grained occupancy in indoor 3D space and propose a 3D occupancy world model for robots to achieve this.
    
\subsection{Occupancy World models}
    Understanding the scene and forecasting the evolutions enable the embodied agents to plan ahead and make decisions in exploration due to their ability to imagine the future. While previous methods \cite{salzmann2020trajectron++, giuliari2021transformer, ngiam2021scene, zhou2022hivt, tang2024hpnet} simulate scene evolutions through trajectory prediction of potential instances, current works use the occupancy world model as a generative framework for describing fine-grained overall scene dynamics. OccWorld \cite{zheng2024occworld} utilizes an autoregressive transformer with a scene tokenizer \cite{van2017neural} to predict future fine-grained occupancy based on historical information alone. OccSora \cite{wang2024occsora} proposes a 4D scene encoder to produce a compact spatiotemporal representation and uses a diffusion model to generate 4D occupancy sequences based on given trajectory prompts. However, current methods \cite{zhang2024efficient, wei2024occllama, yang2025driving, yan2024renderworld, xu2025occ, gu2024dome} focus on outdoor scenes and ignore the exploration of 3D occupancy scene evolutions for robots in indoor scenes. In this work, we restructure a new benchmark for indoor 3D occupancy scene evolution prediction task and explore a new framework for forecasting the future occupancy in indoor 3D space based on the occupancy world model. 

\section{Method}

\subsection{Problem Formulation}

    In this work, we aim to forecast the future 3D occupancy from current 3D occupancy obtained by indoor monocular RGB image and the history occupancy frames. The current 3D occupancy prediction branch is as shown in below:
    \begin{equation}
    O_{current} = F_{mono}(I_{rgb}, E, K)
    \label{eq1}
    \end{equation}
    
    Formally, we are given the RGB image $I_{rgb} \in \mathbb{R}^{H \times W \times 3}$ from the indoor monocular camera, where the $\{H,W\}$ denotes the image resolution. The extrinsic matrix $E\in \mathbb{R}^{3 \times 4}$ and intrinsic matrix $K\in \mathbb{R}^{3 \times 3}$ can be obtained via camera calibration. The $F_{mono}$ is the monocular occupancy prediction model (e.g. EmbodiedOcc \cite{wu2024embodiedocc} and RoboOcc \cite{zhang2025roboocc}), and the $O_{current} \in \mathbb{R}^{X \times Y \times Z \times C_{class}}$ is the current occupancy prediction, where the $\{X,Y,Z\}$ and $C_{class}$ denote the target volume resolution of the front view and the set of semantic classes.

    When we obtain the current occupancy and intend to move to the next viewpoint, the future occupancy forecasting branch is as shown in below:
    \begin{equation}
    O_{future} = F_{world}(O_{current}, O_{history}, P)
    \label{eq2}
    \end{equation}

    Given the current occupancy $O_{current}$, history occupancy $O_{history} \in \mathbb{R}^{T \times X \times Y \times Z \times C_{class}}$ of past observations and the next camera pose $P$ comes from the robot's decision, we obtain the future occupancy forecast $O_{future} \in \mathbb{R}^{X \times Y \times Z \times C_{class}}$ via the occupancy world model $F_{world}$, where the $T$ denotes the number of history frames.

    \begin{figure*}[ht]
    \centering
        \includegraphics[width=1\textwidth]{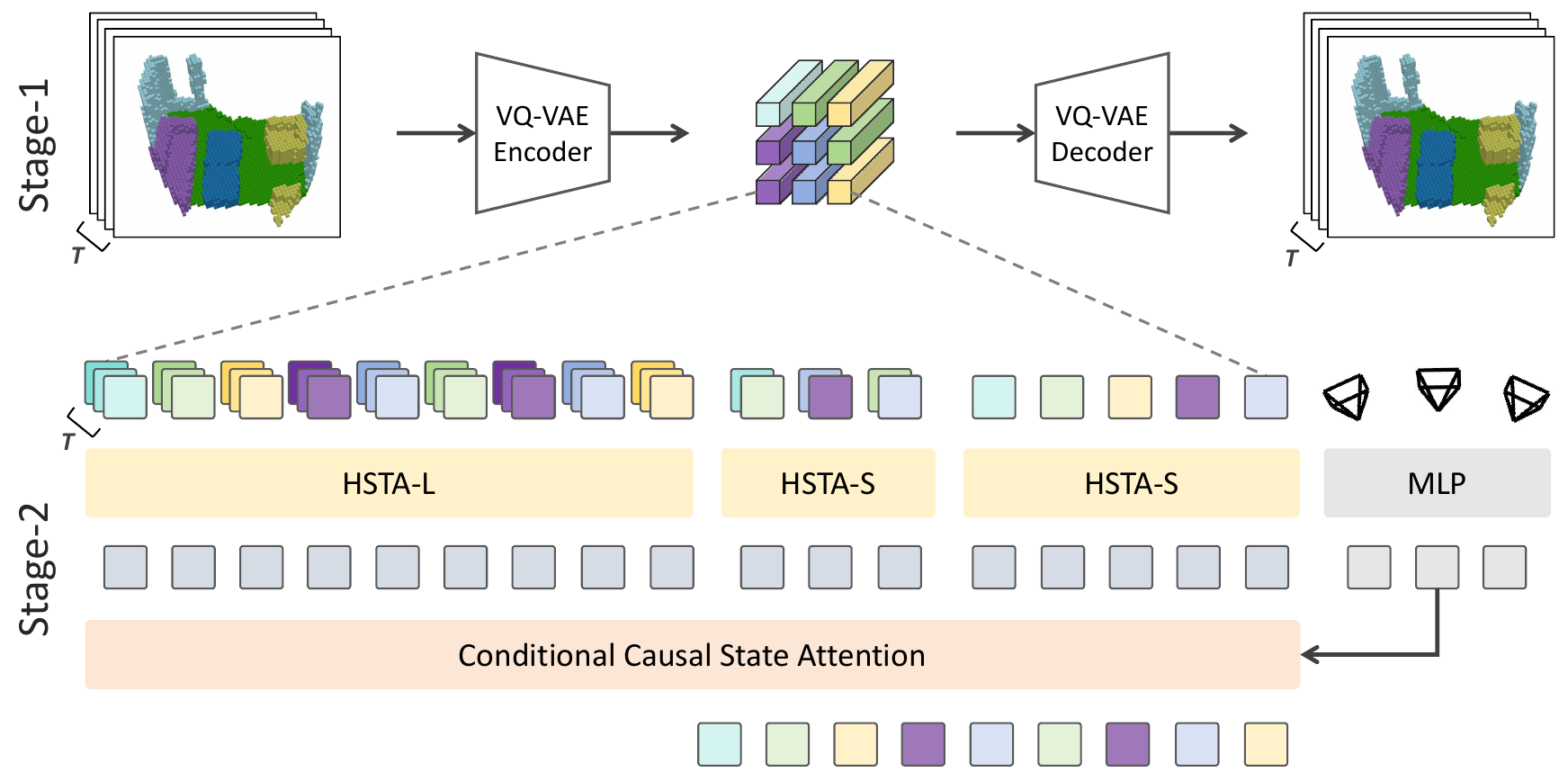}
        \caption{\textbf{The Framework of RoboOccWorld.} The RoboOccWorld is divided into 2 stages. In the first stage, given the current occupancy $O_{current}$, we employ the VQ-VAE as the tokenizer for occupancy scene representation. In the second stage, we partition the scene tokens into multi-scale spatio-temporal chunks, and utilize the Hybrid Spatio-Temporal Aggregation (HSTA) module that consists of both Long-term (L) and Short-term (S) contexts to effectively aggregate the spatio-temporal information. Subsequently, the Conditional Causal State Attention (CCSA) mechanism utilizes an autoregressive transformer guided by the next-step camera pose to generate forecasts for the next state occupancy. }
        \label{fig_4}
    \end{figure*}

\subsection{RoboOccWorld}
    \textbf{Overall Architecture.} As shown in Figure \ref{fig_4}, the overall RoboOccWorld framework consists of 2 stages. In stage 1, given the current occupancy $O_{current}$, we train the VQ-VAE \cite{van2017neural} as the occupancy scene tokenizer. The current occupancy is encoded to the discrete tokens $O_{Z}$, and decoded to reconstruct the vanilla occupancy. In stage 2, we partition the scene tokens into multi-scale spatio-temporal chunks, and utilize the Hybrid Spatio-Temporal Aggregation (HSTA) module that consists of both Long-term (L) and Short-term (S) contexts to effectively expand the spatio-temporal receptive field and aggregate spatio-temporal information. Then the proposed Conditional Causal State Attention (CCSA) use the autoregressive transformer guided by next-step camera pose to make next state occupancy forecast.

    \textbf{Occupancy Scene Tokenizer.} We compress the occupancy into a Bird's-Eye View (BEV) representation in the height dimension by assigning each category with a learnable class embedding and concatenating them in the height dimension. The encoder which consists of 2D convolution layers is used to extract the high-dimensional information of the BEV representation and encode the information into discrete vectors using a learnable codebook. We then utilize the decoder to reconstruct the BEV representation from the learned discrete scene representation $O_{Z}$.

    \begin{wrapfigure}{l}{0.3\textwidth}
    \centering
    \vspace{-2mm}
    \includegraphics[width=\linewidth]{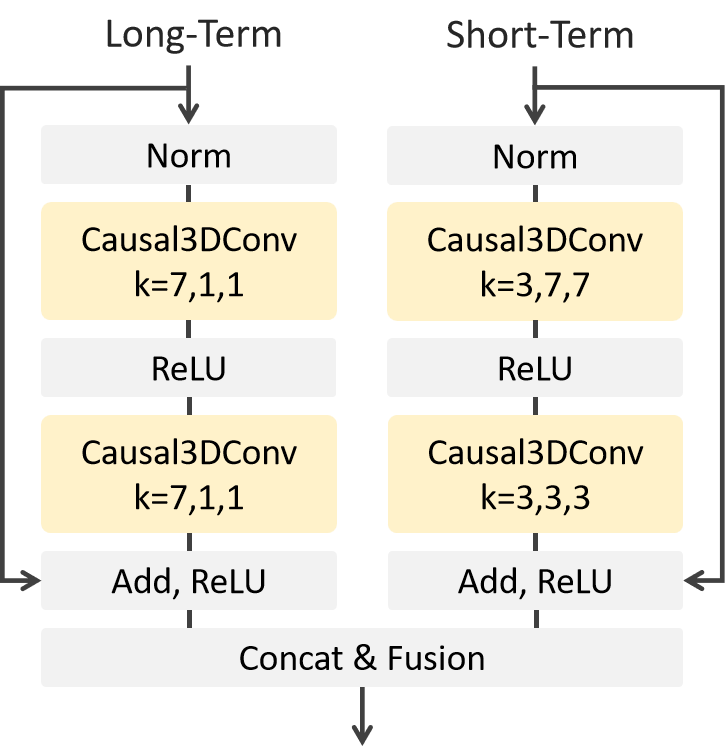}
    \caption{The pipeline of the proposed HSTA.}
    \label{fig_5}
    \vspace{-5mm}
    \end{wrapfigure}
    
    \textbf{Hybrid Spatio-temporal Aggregation.}
    Relying only on spatial aggregation within frames and grid-to-grid spatio-temporal interaction between frames is not enough. Therefore, we propose multi-scale spatio-temporal interactions based on causal 3D convolution to obtain the combined spatio-temporal receptive field. Specifically, we divide spatio-temporal aggregation into two parts: the long-term window and the short-term window. Long-term windows are used to capture the global consistency of the temporal sequence, while short-term windows are used to obtain fine-grained local spatio-temporal information, as shown in Figure \ref{fig_5}. Given the discrete tokens $O_{Z} \in \mathbb{R}^{C \times T \times X \times Y}$, the long-term is as shown in below:
    \begin{align}
    O_{L1} &= ReLU(Causal3DConv1(Norm(O_{Z})))\\
    O_{L2} &= ReLU(Causal3DConv2(O_{L1})+O_{Z})
    \label{eq3}
    \end{align}

    Where the $Causal3DConv_{L1}$ and $Causal3DConv_{L2}$ denote the causal 3D convolution layers with kernel size 7 in temporal dimension and 1 in the spatial dimension. The $Norm$ denotes the BatchNorm3D operator. The long-term windows allow us to obtain historical information for longer timeseries and obtain the global temporal consistency. On the other hand, the short-term is as shown in below:
    \begin{align}
    O_{S1} &= ReLU(Causal3DConv_{S1}(Norm(O_{Z})))\\
    O_{S2} &= ReLU(Causal3DConv_{S2}(O_{S1})+O_{Z})
    \label{eq5}
    \end{align}

    \begin{wrapfigure}{r}{0.3\textwidth}
    \centering
    \vspace{-2mm}
    \includegraphics[width=\linewidth]{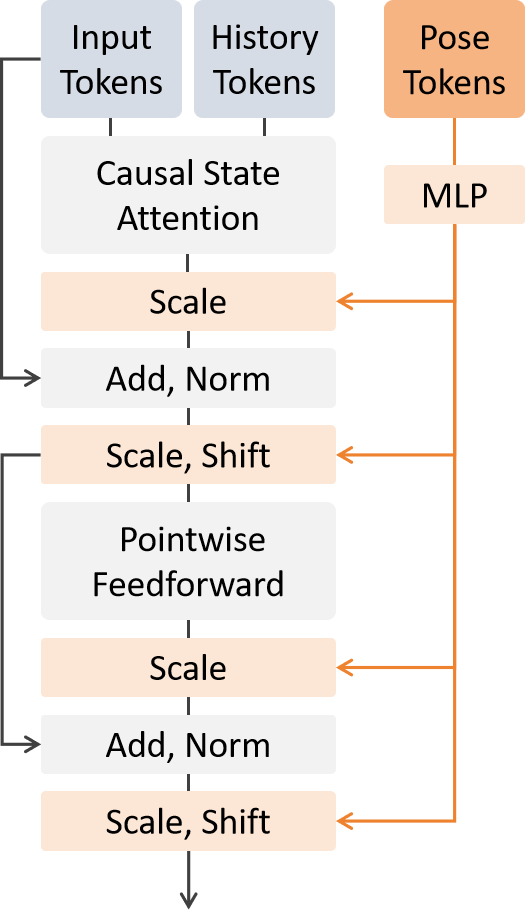}
    \caption{The pipeline of the proposed CCSA.}
    \label{fig_6}
    \vspace{-5mm}
    \end{wrapfigure}
    
    Where the $Causal3DConv_{S1}$ and $Causal3DConv_{S2}$ denote the causal 3D convolution layers with kernel size 3 in temporal dimension and 7 or 3 in the spatial dimension. The short-term windows allow us to obtain the larger spatial receptive field, further capturing local spatio-temporal coherence and fine-grained spatio-temporal information.

    We then concatenate and fuse the spatio-temporal features of the long-term and short-term windows in the channel dimension. With the proposed multi-scale spatio-temporal window, we effectively maintain the temporal consistency of long sequences and effectively utilize the fine-grained spatio-temporal cues of history frames.
    
    \textbf{Conditional Causal State Attention.} Due to the flexible viewpoint transitions and unconstrained movement of robots in indoor environments, the temporal correlation between successive frames becomes significantly weakened compared to outdoor scenes. The increased diversity in camera positions within indoor scenes not only disrupts the inherent spatio-temporal coherence of historical data but also poses significant challenges for autoregressive prediction frameworks that rely on sequential dependency. Therefore, we propose to use the next pose of the camera as conditions to bootstrap the autoregressive transformer for reasonable occupancy prediction, as shown in Figure \ref{fig_6}. Given the discrete tokens $O_{Z}$, input tokens $O_{I}$ and history tokens $O_{H} \in \mathbb{R}^{X \times Y \times T \times C}$ are obtain by causal embeddings. And the camera pose tokens $T_{P}$ is encoded by 2 linear layers from camera pose $P$. We first obtain the scale and shift parameters as conditions from camera pose tokens, the pipeline is as shown in below:
    \begin{equation}
    \alpha_1, \beta_1, \gamma_1, \alpha_2, \beta_2, \gamma_2 = MLP(T_{P})
    \label{eq7}
    \end{equation}

    Where the $\{\alpha, \beta, \gamma\}$ denote the scale factor of residuals, scale factor and shift factor of layer normalization, respectively. The $MLP$ consists of 1 linear layer. Then, the attention branch is as shown in below:
    \begin{align}
    O_{I1} &= (\alpha_1 \cdot CSA(O_{I}, O_{H}, O_{H})) + O_{I}\\
    O_{I1} &= Norm(O_{I1}) \cdot (1 + \beta_1) + \gamma_1
    \label{eq8}
    \end{align}

    The $CSA$ denotes the Causal State Attention, which is specifically the inter-frame cross attention operator with the causal attention mask. The $\{O_{I}, O_{H}, O_{H}\}$ are $query$, $key$, $value$ of the cross attention, respectively. Subsequently, the feedforward branch is as shown in below:
    \begin{align}
    O_{I2} &= (\alpha_2 \cdot FFN(O_{I1})) + O_{I1}\\
    O_{I2} &= Norm(O_{I2}) \cdot (1 + \beta_2) + \gamma_2
    \label{eq10}
    \end{align}

    The $FFN$ consists of 2 linear layers. With the proposed conditional causal state attention, we utilize the visual observations and potential camera poses to guide the generative occupancy world model to make reasonable occupancy forecasts in indoor scenes.

    \textbf{Loss Functions.} The RoboOccWorld is trained in 2 stages following OccWorld \cite{zheng2024occworld}. In the first stage, given the current occupancy, we train the VQ-VAE as the tokenizer for occupancy scene representation using cross-entropy loss $L_{ce}$, lovasz-softmax loss $L_{lovasz}$ and embedding loss $L_{embed}$. In the second stage, we use the learned scene tokenizer to obtain scene tokens and constrain the discrepancy between predicted tokens and scene tokens using cross-entropy loss $L_{ce}$. 
    
\section{Experiment}

    In this paper, we propose an OccWorld-ScanNet benchmark based on the Occ-ScanNet \cite{yu2024monocular} and EmbodiedOcc-ScanNet \cite{wu2024embodiedocc} datasets. We conduct comparative experiments on OccWorld-ScanNet dataset with next state occupancy prediction task and autoregressive occupancy prediction task to validate the effectiveness of the proposed occupancy world model. In next state occupancy prediction task, all of our historical information comes from the ground truth occupancy of the historical state. We aggregate historical information to make occupancy predictions for the next state. On the contrary, we add the predicted occupancy to the historical state and then make occupancy predictions for the next state in autoregressive occupancy prediction task.

\subsection{Datasets}

    \textbf{OccWorld-ScanNet} comprises 537 / 137 scenes in the train/val splits and the scenes in the training set are different from those in the validation set. Each scene in the OccWorld-ScanNet dataset consists of 100 posed image frames with their corresponding occupancy. The dataset provides frames with 12 classes including 1 for free space, and 11 for specific semantics (ceiling, floor, wall, window, chair, bed, sofa, table, tvs, furniture, objects). The annotated voxel grid spans a $4.8m\times4.8m\times2.88m$ box in front of the camera with a resolution of $60\times60\times36$.

\subsection{Evaluation Metrics}

    Following common practices \cite{cao2022monoscene, huang2024gaussianformer}, we use mean Intersection-over-Union (mIoU) and Intersection-over-Union (IoU) to evaluate the performance of our model:
    \begin{equation}
    {\rm mIoU} = \frac{1}{{C_{class}}}\sum_{i\in C_{class}} \frac{TP_i}{TP_i+FP_i+FN_i}, \quad {\rm IoU}=\frac{TP_{\neq C_{0}}}{TP_{\neq C_{0}}+FP_{\neq C_{0}}+FN_{\neq C_{0}}}
    \label{eq12}
    \end{equation}
    
    Where $C_{class}$, $C_{0}$, $TP$, $FP$, $FN$ denote the nonempty classes, the empty class, the number of true positive, false positive and false negative predictions, respectively.

\subsection{Implementation Details}

    \textbf{Scene Tokenizer.} We perform 2$\times$ scaling of the feature resolution in the encoder-decoder, and the learnable codebook comprises 512 nodes and 64 channels. For training settings, we utilize the AdamW \cite{loshchilov2017decoupled} optimizer with a weight decay of 0.01. We set an initial learning rate of 1e-3 and decrease it according to a cosine schedule. We train our scene tokenizer for 200 epochs using 8 A100 GPUs on the OccWorld-ScanNet dataset with the first 16 frames of each scene. The scene tokenizer is trained only in the first stage and is frozen in the second stage.

    \textbf{World Model.} The world model is divided into 3 parts, encoder, middle layers and decoder. Each layer of the encoder consists of 1 HSTA layer and 1 CCSA layer stacked. On the contrary, each layer of the decoder consists of 1 CCSA layer and 1 HSTA layer stacked. The encoder and decoder have 6 layers each. The middle layers consist of 2 ResNet blocks \cite{he2016deep}. For training settings, we utilize the AdamW \cite{loshchilov2017decoupled} optimizer with a weight decay of 0.01, an initial learning rate of 1e-3 and a cosine schedule. We train our world model for 200 epochs using 8 A100 GPUs on the OccWorld-ScanNet dataset. 
    
    During training, we randomly pick 17 consecutive frames from 100 frames of each scene as input, where the first 16 frames are used as history frames and the last 16 frames are used as ground truth for prediction frames. For inference, in next state occupancy prediction task, we take the first 17 frames of each scene as input and the rest is the same as the training setup. Instead, in autoregressive occupancy prediction task, we take the first 6 frames of each scene as historical information and make autoregressive predictions for the next 6 frames.

    \begin{table*}[ht]
    \renewcommand{\arraystretch}{1.2}
        \caption{
            \textbf{Next State Occupancy Prediction on the OccWorld-ScanNet Validation Unseen Set.} The state-of-the-art results are marked with \textbf{boldface}.
            }
            \small
    	\setlength{\tabcolsep}{0.008\textwidth}
            \vspace{-2mm}
            \begin{center}
            \resizebox{1.0\linewidth}{!}{
    		\begin{tabular}{l|c|c|c c c c c c c c c c c|c}
    			\toprule
    			Method
    			& Input
    			& Recon
    			& \rotatebox{90}{\parbox{1.5cm}{\textcolor{ceiling}{$\blacksquare$} ceiling}} 
    			& \rotatebox{90}{\textcolor{floor}{$\blacksquare$} floor}
    			& \rotatebox{90}{\textcolor{wall}{$\blacksquare$} wall} 
    			& \rotatebox{90}{\textcolor{window}{$\blacksquare$} window} 
    			& \rotatebox{90}{\textcolor{chair}{$\blacksquare$} chair} 
    			& \rotatebox{90}{\textcolor{bed}{$\blacksquare$} bed} 
    			& \rotatebox{90}{\textcolor{sofa}{$\blacksquare$} sofa} 
    			& \rotatebox{90}{\textcolor{table}{$\blacksquare$} table} 
    			& \rotatebox{90}{\textcolor{tvs}{$\blacksquare$} tvs} 
    			& \rotatebox{90}{\textcolor{furniture}{$\blacksquare$} furniture} 
    			& \rotatebox{90}{\textcolor{objects}{$\blacksquare$} objects} 
    			& IoU / mIoU\\
                \midrule
                OccWorld~\cite{zheng2024occworld} & $Occ$ & 73.39 / 57.82 & 6.57 & 45.15 & 18.61 & 20.39 & 31.54 & 45.18 & 37.18 & 29.81 & \textbf{0.83} & 33.94 & 17.27 & 27.21 / 26.04 \\
                \rowcolor{gray!20}
                RoboOccWorld~(ours) & $Occ$ & 73.39 / 57.82 & \textbf{8.38} & \textbf{58.09} & \textbf{37.98} & \textbf{29.48} & \textbf{50.37} & \textbf{55.00} & \textbf{55.66} & \textbf{43.71} & 0.56 & \textbf{50.52} & \textbf{31.00} & \textbf{49.55} / \textbf{38.25}\\
                \bottomrule 
                \end{tabular}}
                \end{center}
                \vspace{-2mm}
    		  \label{table1}
    \end{table*}	
    
    \begin{table*}[ht]
    \renewcommand{\arraystretch}{1.2}
        \caption{
            \textbf{Autoregressive Occupancy Prediction on the OccWorld-ScanNet Validation Unseen Set.} The state-of-the-art results are marked with \textbf{boldface}.
            }
            \small
    	\setlength{\tabcolsep}{0.008\textwidth}
            \vspace{-2mm}
            \begin{center}
            \resizebox{1.0\linewidth}{!}{
    		\begin{tabular}{l|c|c|cccccc|c}
    			\toprule
                &&& \multicolumn{7}{c}{IoU / mIoU}\\
                Method&Input&Recon& Step 1 & Step 2 & Step 3 & Step 4 & Step 5 & Step 6 & \cellcolor{gray!20}Avg\\
                \midrule
                OccWorld~\cite{zheng2024occworld} & $Occ$ & 73.39 / 57.82 & 27.05 / 25.47 & 16.99 / 15.50 & 12.67 / 10.77 & 10.16 / 8.25 & 9.14 / 7.38 & 8.65 / 6.64 & \cellcolor{gray!20} 12.22 / 11.90\\
                \rowcolor{gray!20}
                RoboOccWorld~(ours) & $Occ$ & 73.39 / 57.82 & \textbf{50.85} / \textbf{39.14} & \textbf{41.90} / \textbf{30.50} & \textbf{34.95} / \textbf{23.53} & \textbf{27.73} / \textbf{19.04} & \textbf{23.53} / \textbf{16.55} & \textbf{21.63} / \textbf{14.75} & \textbf{31.70} / \textbf{23.51}\\
                \bottomrule 
                \end{tabular}}
                \end{center}
                \vspace{-2mm}
                \label{table2}
    \end{table*}
    
    \begin{minipage}[ht]{0.55\textwidth}
    \renewcommand{\arraystretch}{1.2}
    \small
        \centering
        \captionof{table}{
            End-to-End Next State Occupancy Prediction.
            }
        \label{tab:table3}
        \resizebox{1.0\linewidth}{!}{
        \begin{tabular}{l|c|c|c}
    			\toprule
    			Method
    			& Input
    			& Vanilla
                & IoU / mIoU
    			\\
                \midrule
                OccWorld-E & $RGB$ & 42.89 / 31.13 & 27.71 / 16.27 \\
                \rowcolor{gray!20}
                RoboOccWorld-E & $RGB$ & 42.89 / 31.13 & \textbf{35.46} / \textbf{21.08}\\
                \midrule
                OccWorld-R & $RGB$ & 45.78 / 32.55 & 29.76 / 16.82 \\
                \rowcolor{gray!20}
                RoboOccWorld-R & $RGB$ & 45.78 / 32.55 & \textbf{37.79} / \textbf{22.08}\\
                \bottomrule 
                \end{tabular}}
    \end{minipage}
    \hfill
    \begin{minipage}[ht]{0.4\textwidth}
    \renewcommand{\arraystretch}{1.2}
    \small
        \centering
        \captionof{table}{
            Ablation on the Components of RoboOccWorld.
            }
        \label{tab:table4}
        \vspace{-2mm}
        \begin{tabular}{c c|c c}
    			\toprule
    			CCSA
    			& HSTA
    			& IoU
    			& mIoU\\
                \midrule
                - & - & 27.21 & 26.04 \\
                \checkmark & - & 33.78 & 28.38 \\
                - & \checkmark & 27.61 & 26.20 \\
                \rowcolor{gray!20}
                \checkmark & \checkmark & \textbf{49.55} & \textbf{38.25}\\
                \bottomrule 
                \end{tabular}
    \end{minipage}

    \begin{figure*}[!t]
    \vspace{-2mm}
    \centering
        \includegraphics[width=1\textwidth]{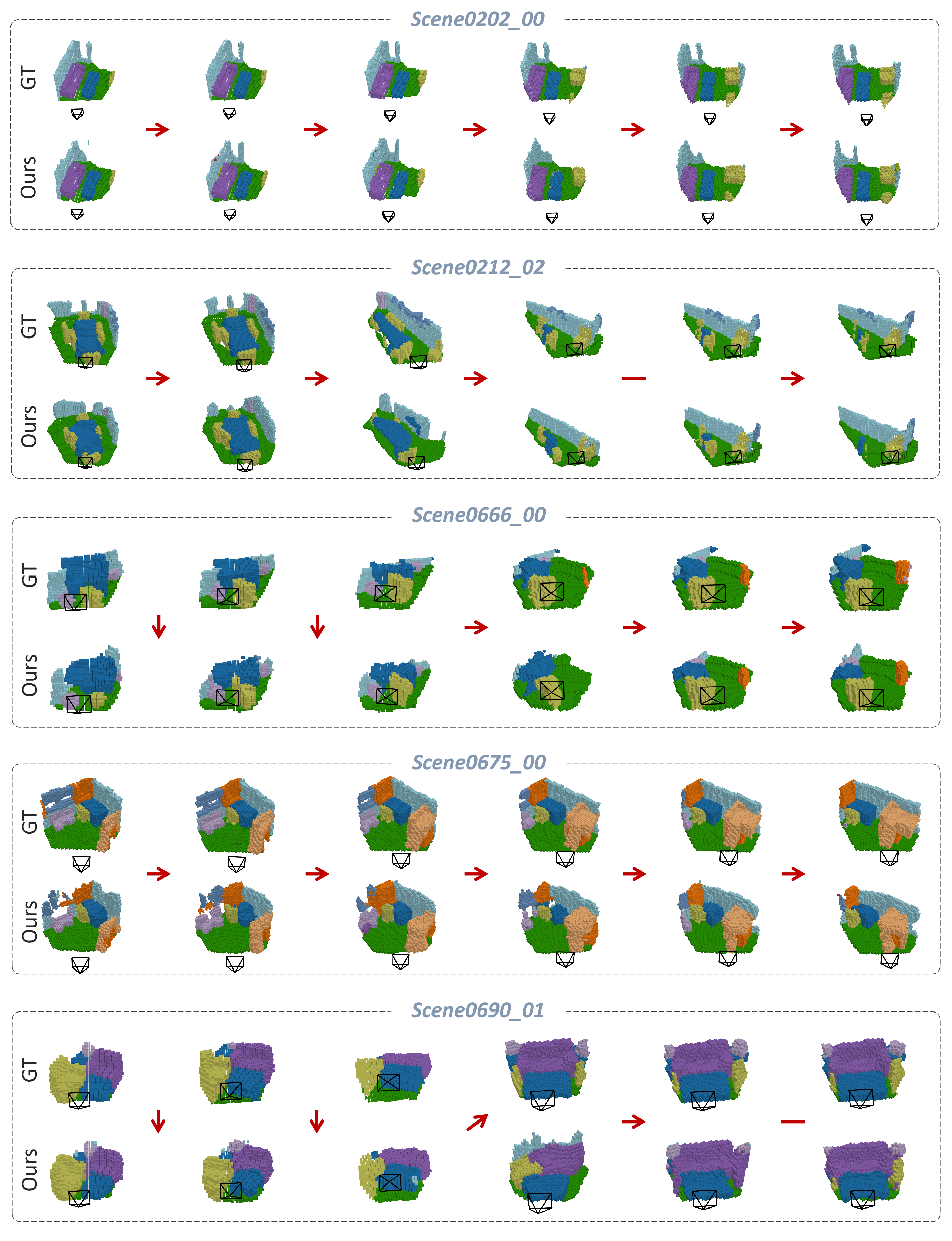}
        \caption{\textbf{Qualitative Analysis on the OccWorld-ScanNet dataset in next state occupancy prediction task.} The \textcolor{redarrows}{red arrows} represent the next pose of the camera. It can be seen that model can successfully understand the pose changes of the camera, which utilizes the pose changes and the spatio-temporal understanding of the scene for accurate prediction of the next state. The proposed method captures the scene layout and forecasts various semantic instances accurately in multiple scenes, which indicates the effectiveness and scalability of the proposed components.}
        \vspace{-5mm}
        \label{fig_7}
    \end{figure*}
    
\subsection{Main Results}

    \textbf{Next State Occupancy Prediction.} We evaluate our model in next state occupancy prediction task, as shown in Table \ref{table1}. In this task, we use 3D occupancy as observations to the world model. Specifically, it is an ideal condition where all observations come from ground truth. The $Recon$ results represent the reconstruction accuracy in IoU and mIoU metric from VQ-VAE \cite{van2017neural}. The proposed RoboOccWorld significantly outperforms the state-of-the-art method by (22.34, 12.21) in IoU and mIoU metric, respectively. Experimental results demonstrate the strengths of combined spatio-temporal receptive field and pose-guided autoregressive transformer. 

    \textbf{Autoregressive Occupancy Prediction.} We then evaluate our model in autoregressive occupancy prediction task, as shown in Table \ref{table2}. In this task, we also use 3D occupancy as observations to the world model. However, ground truth is no longer the only option for the sequence of historical observations we wish to maintain. Specifically, we make the first prediction based on ground truth, and subsequently add the prediction to the historical observation series, and so on, with autoregression to complete the remaining predictions. The $Recon$ results also represent the reconstruction accuracy in IoU and mIoU metric from VQ-VAE \cite{van2017neural}. The proposed RoboOccWorld significantly outperforms the state-of-the-art method by (19.48, 11.61) in IoU and mIoU metric, respectively. It can be seen that our proposed method remains competitive even under the challenging autoregressive prediction task, which demonstrates the potential of the method.

    \textbf{Visualizations.} In Figure \ref{fig_7}, we qualitatively analyze the proposed RoboOccWorld in next state occupancy prediction task. It can be seen that model can successfully understand the pose changes of the camera. It utilizes the pose changes and the spatio-temporal understanding of the scene for accurate prediction of the next state. The proposed method captures the scene layout and forecasts various semantic instances accurately in multiple scenes, which indicates the effectiveness and scalability of the proposed components.

\subsection{Ablation Study}

    \textbf{Analysis on the End-to-End Next State Occupancy Prediction.} In Table \ref{tab:table3}, we employ the EmbodiedOcc \cite{wu2024embodiedocc} and RoboOcc \cite{zhang2025roboocc} as the competitive indoor monocular occupancy prediction model for end-to-end next-state prediction. In this task, we use monocular rgb images as observations to the world model. Specifically, the monocular occupancy model receives monocular rgb images and predicts fine-grained semantic occupancy of the current scene. Based on the current and historical occupancy obtained from the monocular model, we utilize the world model to make occupancy predictions for the next state. The $E$ and $R$ represent the EmbodiedOcc and RoboOcc. The $Vanilla$ represents the IoU and mIoU of the predictions from image to occupancy in different indoor monocular occupancy prediction model. It can be seen that our RoboOccWorld not only substantially outperforms the OccWorld, but also remains competitive for different occupancy predictions as inputs, demonstrating its effectiveness and scalability.
    
    \textbf{Analysis on the Components of RoboOccWorld.} In Table \ref{tab:table4}, we provide analysis on the components of our proposed method in next state occupancy prediction task to validate the effectiveness. What can be seen is that the proposed CCSA brings a performance increase for world model due to its guidance to generative autoregressive transformer. Benefiting from visual observations and potential camera poses, the occupancy world model can make more reasonable forecasts. For the proposed HSTA, expanding the spatio-temporal receptive field only produces only slight changes as the network still lacks the ability to capture spatio-temporal regularity. However, when we activate all the proposed components at the same time, a qualitative change in performance occurs. With the guidance of camera poses and the combined spatio-temporal receptive field, the network captures multi-scale spatio-temporal information and makes reasonable occupancy predictions for the next state, which demonstrates that the world model starts to understand the spatio-temporal regularity in exploration.
    
\section{Conclusion}

    Based on the understanding of the scene and the forecasting of its evolution, robots can perform better exploration and make better decisions. Currently, occupancy world models have received increasing attention as a generative framework as they can be used to describe fine-grained scene evolution rather than instance-level trajectory prediction. However, occupancy scene evolution prediction for robots in indoor scenes is still not effectively explored. In this paper, we restructure the OccWorld-ScanNet benchmark to evaluate the forecasts of scene evolutions and propose a occupancy world model based on the combined spatio-temporal receptive field and guided autoregressive transformer for robots' decision and exploration. Both quantitative and qualitative results have shown that our RoboOccWorld could effectively forecast future scene evolutions in the comprehensive 3D semantic occupancy space and outperforms the state-of-the-art methods. We hope our work can shed light on world model for robots.

    \textbf{Limitations.} Although our proposed world model could effectively forecast future fine-grained occupancy, it focuses only on the occupancy representation of the 3D scene while ignoring other representations. Image is an intuitive and more challenging representation in generative tasks. Therefore, forecasting for future images should be explored in subsequent work. In addition, the current work focuses on a single perception task and aims to provide future scene-level predictions for the perception module, without serving subsequent modules (e.g., navigation tasks). Therefore, expanding the world model to navigation tasks should also be considered in subsequent work.
    
\clearpage
\bibliographystyle{unsrt}
\bibliography{neurips_2025.bib}

\clearpage

\end{document}